\documentclass[journal]{IEEEtran}
\usepackage{amsmath,amsfonts}
\usepackage{algorithmic}
\usepackage{algorithm}
\usepackage{array}
\usepackage{textcomp}
\usepackage{stfloats}
\usepackage{url}
\usepackage{verbatim}
\usepackage{graphicx}
\usepackage{cite}
\usepackage{subfigure} 
\usepackage{booktabs}
\usepackage{array}

\usepackage{placeins}
\usepackage{dsfont}
\usepackage{amsmath,amssymb,amsfonts}
\usepackage{algorithm}
\usepackage{makecell}

\usepackage{amsthm}
\theoremstyle{definition} 

\usepackage{color}
\usepackage{breqn}
\usepackage{enumitem}

\usepackage[absolute]{textpos}
\usepackage{lipsum}

\usepackage{pifont}

\usepackage{newfloat}
\usepackage{listings}
\usepackage{multirow}

\begin{document}

\title{Towards Faithful Graph Explanations with Synergistic Edge Effects via Granular Balls}
%
\author{Jiancu Chen, Shuyin Xia\textsuperscript{\rm *}, \textit{Senior Member, IEEE}, Guan Wang, Degang Chen, Fan Chen 

\thanks{Manuscript received April xx, xxxx; revised August xx, xxxx. This research was partially supported by the National Natural Science Foundation of China under Grant Nos. 62222601, 62176033, 62221005, and 61936001; the Chongqing Municipal Education Commission Youth Project No. KJQN202301231; the Key Cooperation Project of the Chongqing Municipal Education Commission under Grant No. HZ2021008; the Scientific and Technological Research Program of the Chongqing Municipal Education Commission under Grant No. KJZD-M202400603; and the Project of the Key Laboratory of Tourism Multisource Data Perception and Decision-Making, Ministry of Culture and Tourism, under Grant No. H2023009.
(*Corresponding author: Shuyin Xia).}%
\thanks{
Jiancu Chen, Shuyin Xia, Guan Wang, Degang Chen, and Fan Chen are with the Chongqing Key Laboratory of Computational Intelligence, Key Laboratory of Big Data Intelligent Computing, Key Laboratory of Cyberspace Big Data Intelligent Security, Ministry of Education, School of Computer Science and Technology, Chongqing University of Posts and Telecommunications, Chongqing, China. Jiancu Chen is also with the School of Computer Science and Engineering, Chongqing Three Gorges University, Chongqing, China. (e-mail: chenchen2153@163.com, xiasy@cqupt.edu.cn, 1640124287@qq.com, chendegang0204@163.com, d240201002@stu.cqupt.edu.cn).
}
}


\IEEEpubid{}

\maketitle

\begin{abstract}
Instance-level explanations aim to reveal the rationale behind a model's decisions for a specific graph. Previous methods explain graph neural networks (GNNs) by selecting important edges to induce subgraphs, where edge importance is assessed by perturbing each edge and observing changes in the model predictions. However, they often neglect the synergistic effects among edges, which are crucial for accurately characterizing edge importance. To address this issue, we propose \textbf{SeeExplainer}, a parameter-free explainer to interpret GNNs. Specifically, we first introduce a granular-ball graph refinement mechanism that decomposes a graph into several disjoint granular-balls with no fixed size, and utilize them as nodes to construct a structural graph. This process can better capture the synergistic effects among edges. Then, we perturb nodes and edges in the structural graph to generate explanatory subgraphs based on their respective contributions. Experiments on several graph classification datasets of different networks show that SeeExplainer outperforms state-of-the-art baselines. 
\end{abstract}

\section{Introduction}
\label{sec_introduction}
\IEEEPARstart{I}{n} recent years, Graph Neural Networks (GNNs) have attracted significant attention for their outstanding performance in graph classification tasks\cite{xie2022active, xie2022semisupervised, cheng2025edge}. Since graph data is widely found in real-world domains, such as social networks\cite{cai2018simple}, chemistry\cite{morris2020tudataset}, and biology\cite{borgwardt2005protein, sutherland2003spline}, the practical value of GNNs has become increasingly prominent. However, like many deep learning models, the ``black-box" nature of GNNs makes their internal mechanisms difficult to understand. Without interpretation of the model’s underlying mechanisms, it is hard to establish trust, which directly affects its application in critical areas such as fairness, privacy protection, and security \cite{doshi2017towards}. Therefore, explaining GNNs has become a key research area in explainable machine learning \cite{yuan2022explainability, ying2019gnnexplainer}.

\begin{figure}[H]
    \centering
    \includegraphics[width=8cm,height=5.165cm]{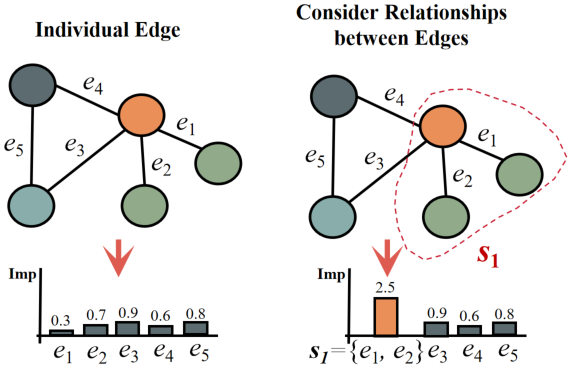}
	\caption{The synergistic effects among edges. A substructure $s_1$ exhibits a significantly higher contribution than the sum of its individual edges $e_1$ and $e_2$.}
	\label{motivation}
\end{figure}

To interpret graph neural networks, researchers have proposed various explanation techniques. Model-level explanations\cite{yuan2020xgnn,wang2022gnninterpreter,chen2023d4explainer,wang2024gnnboundary,saha2024graphon} aim to reveal the global structural patterns and decision strategies captured by the model. Instance-level explanations aim to reveal the reasons behind a model’s decisions on specific graph instances. Unlike model-level explanations, instance-level explanations offer more fine-grained and localized insights, aligning more closely with human reasoning and scientific research. Early methods primarily focused on node contributions, that is, revealing which entities contribute to the model’s predictions. These methods typically highlight the most representative regions of the graph by learning node masks or scoring functions. GNNExplainer \cite{ying2019gnnexplainer} jointly optimizes node masks with the prediction target. Perturbation-based strategies assess node contributions by removing or perturbing nodes. GraphLime \cite{huang2022graphlime} observes changes in model predictions through local node perturbations to identify key nodes that affect classification decisions. 

These methods rely on node information but overlook the fact that selecting an edge naturally entails selecting the corresponding endpoints \cite{lu2024eigsearchgeneratingedgeinducedsubgraphs}, making it difficult to effectively explore subgraph-level explanations. To address this issue, recent methods have begun to focus on edge contributions. GraphMask \cite{schlichtkrull2020interpreting} maintains the minimal set of edges that preserve the prediction results through sparsity constraints. SubgraphX \cite{yuan2021explainability} evaluates edge contributions to provide more comprehensive explanations. PGExplainer \cite{NEURIPS2020_e37b08dd} models edges by learning a parameterized edge mask generator. Eig-Search \cite{lu2024eigsearchgeneratingedgeinducedsubgraphs} generates induced subgraphs by selecting important edges, and uses them to generate subgraph-level explanations. 

Nevertheless, the aforementioned methods emphasize the importance by perturbing an individual edge, but overlook the synergistic effect among edges, which hampers the accurate identification of edge importance. For example, in graph coloring problems \cite{zhang2006introduction}, edges sharing vertices cannot have the same color; in network flow problems\cite{waissi1994network}, edges that share vertices may be subject to shared traffic constraints. As shown in Fig. \ref{motivation}(a), multiple edges collectively form a special nitro functional group ($e_1$ and $e_2$), if we split the nitro group and calculate its contribution separately, it loses its specific meaning. As shown in Fig. \ref{motivation}(b), if we calculate the importance of nitro groups as a whole, their contribution is not simply the sum of their independent edge contributions, but rather exhibits nonlinear growth.

To address this issue, we propose SeeExplainer, a graph neural network explainer that captures the synergistic effect among edges and directly generates subgraphs to explain GNN predictions. Inspired by granular-ball computing theory \cite{osti_10286756, 9911987} and the human cognitive concept of ``global precedence", we cover the entire graph using a coarse-grained granular-ball, then obtain a structural graph through non-isomorphic factorization, and perturb nodes and edges in the structural graph to generate an explanatory subgraph. Specifically, we first use a coarse splitting strategy to efficiently explore a coarse-grained representation of the original graph, then use a fine splitting strategy to progressively refine the granular-balls into different sizes and optimal granularities, and construct a structural graph based on these granular-balls. During the construction of the structural graph, the original nodes and edges are not removed; rather, the internal structure of a granular-ball is represented as a whole within the structural graph. This preserves the original graph structure while capturing the synergistic effect among edges, resulting in better explanations. In summary, our contributions are as follows:
\begin{itemize}
    \item \textbf{General Aspects:} We propose a structural graph generation strategy that captures the synergistic effect among edges, offering a new perspective compared with traditional methods that only consider the contribution of individual edges.
    \item \textbf{Novel Methodologies:} We introduce SeeExplainer, a parameter-free explainer to interpret graph neural networks, which directly generates multi-granularity explanatory subgraphs based on the structural graph. Compared with traditional methods that generate induced subgraphs based on important edges, our method is simple and effective.
    \item \textbf{Multifaceted Experiments:} Extensive experiments on public real-world graph classification datasets of different networks demonstrate the efficacy of SeeExplainer, greatly enhancing the explainability of GNNs.
\end{itemize}

\section{Related Work}
\label{sec_related_work}
\subsection{Graph Neural Network}
Graph data is widely used in many real-world domains\cite{sutherland2003spline, vincent2021online, morris2020tudataset}. Unlike image and text data, a graph is represented by a feature matrix and an adjacency matrix. Specifically, consider a graph $G = (V, X, A)$, where $V = \{v_1, v_2, ..., v_n\}$ is the set of nodes, $X\in \mathds{R}^{n \times d} $ is the matrix consisting of the $d$-dimensional feature vectors of all nodes, and $A \in {\{0, 1\}}^{n \times n}$ is the adjacency matrix of $G$. Graph neural networks learn node representations based on these matrices. Although there are various types of graph neural networks, such as Graph Convolutional Networks (GCNs) \cite{DBLP:journals/corr/KipfW16}, and Graph Isomorphism Networks (GINs) \cite{xu2019powerfulgraphneuralnetworks}, they all follow an information aggregation scheme, in which a node’s representation is obtained by aggregating and combining the features of its neighboring nodes. Here, we use GCNs as an example to illustrate the information aggregation scheme. For Graph $G$, the aggregation operation in GCNs can be mathematically written as $X_{i+1}=\sigma (D^{-\frac{1}{2}}\hat{A}D^{-\frac{1}{2}} X_i W_i)$, where $X_i$ denotes the output feature matrix of the $i$-th GCN layer and $X_0$ is set to $X$. The node features are transformed from $X_i \in \mathcal{R}^{n \times c_{i+1}}$. Note that $\hat{A} = A + I$ is employed to add self-loops, and $D$ is a diagonal node degree matrix to perform normalization on $\hat{A}$. In addition, $W_i \in \mathcal{R}^{c_i \times c_{i+1}}$ is a learnable weight matrix to perform linear transformations on features, and $\sigma(\cdot)$ is the non-linear activation function.

\subsection{Instance-level explanations}
Instance-level explanation methods aim to identify the subset of input features that most strongly influence the prediction, providing input-dependent explanations for each graph \cite{yuan2022explainability}. Based on how feature-contribution scores are obtained, instance-level explanation methods can be categorized into gradient- or feature-based methods \cite{baldassarre2019explainability, pope2019explainability}, decomposition-based methods \cite{feng2023degree}, surrogate-based methods \cite{huang2022graphlime, duval2021graphsvx, pereira2023distill, zhang2021relex}, and perturbation-based methods \cite{ying2019gnnexplainer, NEURIPS2020_e37b08dd, wang2021towards, funke2022zorro, yuan2021explainability, zhang2022gstarx}. Perturbation-based methods are widely used to explain deep graph models. Their core motivation is to study how the output changes under different input perturbations. When important features are preserved (i.e., not perturbed), the prediction should remain similar to the original prediction. In practice, these methods employ various mask-generation algorithms to produce different types of masks. Combining a mask with the input graph produces a new graph that retains important input information. This new graph is then fed into the trained GNN to evaluate the mask and update the mask-generation algorithm.

\begin{figure*}[ht]
    \centering  
    \includegraphics[width=17.875cm,height=12cm]{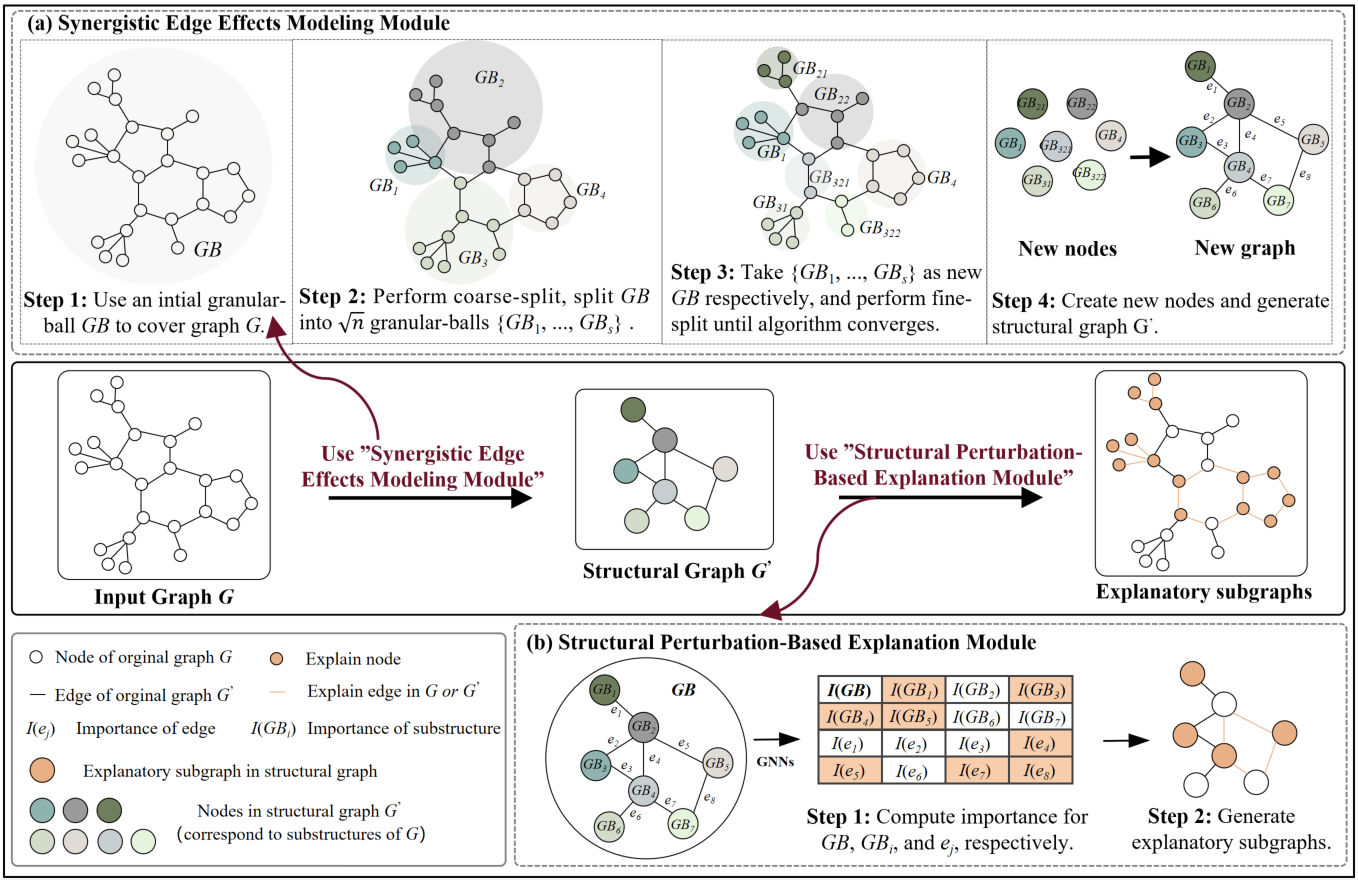}
    \caption{Overview of SeeExplainer. (a) Synergistic Edge Effects Modeling Module: it constructs a structural graph by leveraging information from the original graph and granular-ball computing, aiming to capture the synergistic effects among edges. (b) Structural Perturbation-Based Explanation Module: it employs graph neural networks to quantify the contributions of nodes (i.e., substructures in the original graph) and edges within the structural graph, and subsequently generates explanatory subgraphs.} 
\label{graphnetwork}
\end{figure*}

\subsection{Granular-ball Computing}
In 1982, Chen \cite{chen1982topological} points out that the brain gives priority to recognizing a ``wide range" of contour information in image recognition, and that human cognition exhibits the characteristic of ``global precedence". The human brain's global precedence cognition is efficient and robust, which is beneficial for improving the performance of existing artificial intelligence algorithms. Wang \cite{wang2017dgcc} first introduces the large-scale cognitive rule into granular computing and proposes multi-granular cognitive computing. Xia and Wang \cite{osti_10286756, 9911987} further use granular-balls of different sizes to represent ``grains" and propose granular-ball computing, in which a large granular-ball represents a coarse-grained representation, while a small granular-ball represents a fine-grained representation. Granular-ball computing is initially used to deal with classification problems successfully, and it is also applied to many other learning methods to improve their generalizability or efficiency, such as machine learning \cite{sajid2025gb}, clustering \cite{9139397, xia2025gbct, jia2025generation, li2025granular, xie2025aw}, rough sets \cite{10313988, yang2025constructing}, outlier detection \cite{gao2025fuzzy}, and graph coarsening \cite{xia2025graph}.

\section{Preliminaries}
\subsection{Problem Formulation}
Let $G=(V,E)$ be a graph, where $V=\{v_1, v_2, ...,v_n \}$ denotes the set of nodes and $E \subseteq V \times V$ denotes the set of edges. The adjacency matrix of $G$ is denoted by $A \in \{0, 1\}^{n \times n}$, where $A_{ij} = 1$ if there exists an edge between nodes $v_i$ and $v_j$. Given a pre-trained graph neural network $\phi(\cdot)$ that denotes the logits predicted by the model for the true label $Y$, our work aims to identify the explanatory subgraphs within $G$ that are most influential for the model's prediction. Directly interpreting GNN predictions on the original graph, however, often fails to capture potential synergistic effects among edges. Therefore, we construct a structural graph $G' = (V', E')$ by abstracting the local patterns with synergistic effects in $G$ as structural units. Each node $v' \in V'$ in the structural graph corresponds to a structural unit in the original graph, while edges $(v_i', v_j') \in E'$ encode relationships between these structural units. Based on this abstraction, we formulate the explanation task as identifying a subgraph $\tilde{G} = (\tilde{V}, \tilde{E})$ of the structural graph $G'$, where $\tilde{V} \subseteq V'$ and $\tilde{E} \subseteq E'$. The resulting explanatory subgraphs $\tilde{G}$ should ensure that when the original graph substructure corresponding to $G'$ is used as input, the GNN’s prediction remains faithful to the original output.

\section{Method}
\label{sec_method}
In this section, we propose SeeExplainer, a graph neural network explainer that considers the synergistic effects among edges and generates subgraphs to explain GNN predictions. As shown in Fig. \ref{graphnetwork}, SeeExplainer includes a synergistic edge effects modeling module (in section \ref{4.2}) and a structural perturbation-based explanation module (in section \ref{4.3}).

\subsection{Synergistic Edge Effects Modeling Module}
\label{4.2}
In this module, we aim to generate a structural graph $G^'$ that captures the synergistic effects among edges in an original graph $G$. Firstly, at the beginning of the graph structuring process, we use a granular-ball $GB$ to cover $G$ and perform global computation on the graph, thereby simulating the human cognitive characteristic of ``global precedence". Secondly, we perform non-isomorphic factorization on the graph, dividing it into several disjoint subgraphs with no fixed size. This process can better generates substructures with synergistic effects and utilize them as nodes in the structural graph. Specifically, we perform a coarse-splitting strategy on $GB$, followed by a fine-splitting strategy on each child-ball in $GB$ until the algorithm converges.

\subsubsection{Graph representation and global computation}
Given a graph $G=(V,E)$ with $n$ nodes and $m$ edges, the adjacency matrix of $G$ is denoted by $A \in \mathbb{R}^{n \times n}$, where
\begin{equation}
    A_{ij}=\begin{cases}
  & 1, \text{ if } \left ( i,j \right ) \in E, \\
  & 0, \text{ otherwise }.
\end{cases}
\label{equ1}
\end{equation}
First, we consider the entire graph as a whole and use an initial granular-ball $GB$ to cover and represent graph $G$. Mathematically,
\begin{equation}
    GB = (V, E),
\label{equ3}
\end{equation}
where $GB$ represents the coarsest granularity of $G$.

Then, we compute the degree of each node in $G$. This step is regarded as a global computation and helps accelerate the subsequent granular-ball splitting process. The degree of node $v_i$ is
\begin{equation}
    deg(v_i)=\sum_{j=1}^{n} A_{ij},
\label{equ2}
\end{equation}
where $v_i \in V$. 
\subsubsection{Coarse-splitting strategy} In graph theory, the correlation among edges is generated by sharing vertices\cite{west2001introduction}. This correlation can be formalized by treating each edge as an entity and the shared vertices as interaction channels. Therefore, we select nodes with the highest degrees in $GB$ as center points of the sub-balls. Mathematically,
\begin{equation}
\textit{C} = \left \{ c_1, c_2, \cdots, c_s \right \} =  \operatorname*{argmax} \sum_{v_i \in V} deg(v_i),
\label{equ4}
\end{equation}
where $s=\sqrt{n}$. The selection of $\sqrt{n}$ is empirical, based on previous works \cite{xie2020new, yu2001upper, xia2025graph}. It enables fast processing of large-scale graphs, improving computational efficiency while maintaining a balance between size and quality. 

Then, we use breadth-first search (BFS)\cite{cormen2022introduction} to assign the remaining nodes $v \in V\setminus C$ to obtain child-balls,
\begin{equation}
\mathcal{GB} = \left \{ GB_1, GB_2, \cdots, GB_s \right \}, 
\label{equ5}
\end{equation}
where $GB_1 \cap  GB_2 \cap \dots \cap GB_s = \emptyset$, and $s=\sqrt{n}$.

\subsubsection{Fine-splitting strategy} Next, we perform a fine-splitting strategy on each child-ball $GB_i \in \mathcal{GB}, i \in (1, s)$ respectively. Specifically, treat $GB_i$ as a new parent-ball, and select two nodes with the highest degree in $GB_i$ as the center points. The center points can be represented as, 
\begin{equation}
(c_{i_1},c_{i_2}) = \operatorname*{argmax} \sum_{v \in GB_i} deg(v).
\label{equ6}
\end{equation}
And we use BFS to assign the remaining nodes $v \in GB_i \setminus (c_{i_1},c_{i_2})$ to obtain two clusters $clus\_{i_1}$ and $clus\_{i_2}$.

Then, we calculate the quality of $GB_i$, $clus\_{i_1}$, and $clus\_{i_2}$ respectively. The formula for calculating quality is
\begin{equation}
Q(GB)=\frac{|E_{GB}|}{|V_{GB}|},
\label{equ7}
\end{equation}
where $|E_{GB}|$ indicates the number of edges in $GB$, and $|V_{GB}|$ denotes the number of nodes in $GB$.

Next, we will provide the splitting conditions of a granular-ball.
\begin{equation}
    \begin{cases} 
    Q(clus\_{i_1}) + Q(clus\_{i_2}) > Q(GB_{i})\quad, \quad split; \\ %
    \quad\quad\quad\quad\text{otherwise}\quad\quad\quad \quad\quad\quad,\quad not \quad split.
    \end{cases}
\label{equ8}
\end{equation}

If the splitting conditions are met, the granular-ball $GB_i$ is split into $GB_{i_1}$ and $GB_{i_2}$, where $GB_{i_1}$ corresponds to $clus\_{i_1}$ and $GB_{i_2}$ corresponds to $clus\_{i_2}$. After obtaining $GB_{i_1}$ and $GB_{i_2}$, they are iteratively treated as new parent-balls, and the fine-splitting strategy is applied to them until the algorithm converges. The splitting strategy ensures that all factors in the decomposition process of the graph satisfy a global balance condition. Due to the adaptive number of factors, the overall structure obtained after decomposition also satisfies an inherent balance.

Finally, we consider generated granular-balls as new nodes to generate a structural graph $G^{'} = (V^{'}, E^{'})$ with $p$ nodes and $q$ edges, where $V^{'}=\left \{GB_1, GB_2, \cdots, GB_p \right \}$ and $E^{'} = \left \{e_1, e_2, \cdots, e_q \right \}$. The structural graph $G^{'}$ satisfies $GB_1 \cap  GB_2 \cap \dots GB_p = \emptyset$, and $\bigcup_{i=1}^{p} GB_i \oplus \bigcup_{j=1}^{q} e_j = G$.
The process of a structural graph generation is provided in Algorithm \ref{algorithm1}.

\begin{algorithm}[ht]
    \caption{Generation of a structural graph}
    \textbf{Input}: An original graph $G=(V, X, A)$. \\   
    \textbf{Output}: A structural graph $G^'$.
    \label{algorithm1}
    \begin{algorithmic}[1]    
    \STATE Initialize a granular-ball $GB$ covering $G$, and compute node degrees in $G$ ( Eq. \ref{equ2} );
    \STATE Select $\sqrt{n}$ nodes with the highest degrees as centers, assign remaining nodes by BFS, and split $GB$ into $\sqrt{n}$ granular-balls $\{GB_1,\cdots, GB_s\}$ (Eq. \ref{equ4}-\ref{equ5});
    \STATE Compute the quality $Q(GB_i)$ for each $GB_i$ $(i=1,\cdots,s)$ (Eq. \ref{equ7});    
    \FOR{each parent-ball $GB_i$}
        \STATE Select two highest-degree nodes in $GB_i$ as centers (Eq. \ref{equ6});
        \STATE Split $GB_i$ into two clusters $clus\_{i_1}$ and $clus\_{i_2}$ using BFS;
        \STATE Compute $Q(clus\_{i_1})$ and $Q(clus\_{i_2})$;
        \IF{$Q(clus\_{i_1}) + Q(clus\_{i_2}) > Q(GB_i)$}
            \STATE Split $GB_i$ into two child-balls $GB_{i_1}=clus\_{i_1}$ and $GB_{i_2}=clus\_{i_2}$ and add them to $\mathcal{GB}\_list$;
            \STATE Treat $GB_{i_1}$ and $GB_{i_2}$ as new parent-balls and repeat step 4;
        \ENDIF
    \ENDFOR    
    \RETURN $\mathcal{GB}\_list$
\end{algorithmic}
\end{algorithm}

\subsection{Structural Perturbation-Based Explanation Module}
\label{4.3}
After generating a structural graph $G^{'}$ from $G$, we evaluate it to derive the final explanatory subgraphs. First, we compute the individual contributions of nodes and edges in $G^{'}$, where the contribution of each node in $G^{'}$ corresponds to that of the associated substructure in $G$. Second, we generate explanatory subgraphs based on a simple threshold judgment.

\subsubsection{Contribution assessment}
Given a graph $G=(V,E)$ with $n$ nodes and $m$ edges, its corresponding structural graph is denoted as $G^{'} = (V^{'}, E^{'})$, where $ V^{'}=\left \{ v^{'}_1, v^{'}_2, \cdots, v^{'}_p \right \} = \left \{GB_1, GB_2, \cdots, GB_p \right \}$, and $E^{'} = \left \{e_1, e_2, \cdots, e_q \right \}$. For a graph neural network $\phi(\cdot)$, we adopt the average prediction confidence as a measure of contribution. Thus, the contribution of $G$ is defined as
    \begin{equation}
        I({G}) = \frac{\sum_{i=1}^{m} (\phi (G) - \phi (G \setminus e_i)) }{m}.
    \label{equ9}
    \end{equation}
The contribution of node $v^{'}_i$ in $G^{'}$ is defined as
    \begin{equation}
        I({v^{'}_i}) = \phi (G) - \phi (G \setminus v^{'}_i)),
    \label{equ10}
    \end{equation}
where $v^{'}_i \in V^{'}$, and $i \in (1, p)$.

\noindent The contribution of edge $e^{'}_j$ in $G^{'}$ is defined as
    \begin{equation}
        I({e^{'}_j}) = \phi (G) - \phi (G \setminus e^{'}_j)),
    \label{equ11}
    \end{equation}
where $e^{'}_j \in E^{'}$, and $j \in (1, q)$.

\subsubsection{Explanatory subgraphs generation} When we obtain the contribution of each node and edge in $G^{'}$, we generate an explanatory subgraphs for the original graph $G$. Specifically, we use $I(G)$ as the threshold. If $I(v^{'}_i) > I(G)$, then $v^{'}_i$ is an important substructure in graph $G$. If $I(e^{'}_j) > I(G)$, then $e^{'}_j$ is an important edge in graph $G$. We aggregate $v^{'}_i$ and $e^{'}_j$ that meet the above conditions, which together form the final explanatory subgraphs. 

The explanatory subgraphs are obtained by comparing a threshold with the contribution of local structures or edges. The reason is that the threshold is computed as the average contribution of the entire graph. When the contribution of a local structure or edge exceeds the threshold, it indicates that the local structure has obtained positive synergistic effects among edges during the construction of the structural graph.

The process of explanatory subgraphs generation is provided in Algorithm \ref{algorithm2}.

\begin{algorithm} [ht]
    \caption{Generation of explanatory subgraphs}
    \label{algorithm2}
    \textbf{Input}: A graph neural network $\phi(\cdot)$, a graph $G=(V,E)$ and its structural graph $G^{'}=(V^{'},E^{'})$ where $ V^{'}=\left \{ v^{'}_1, v^{'}_2, \cdots, v^{'}_p \right \} = \left \{GB_1, GB_2, \cdots, GB_p \right \}$ and $E^{'}=\left \{ e_1, e_2, \dots, e_q \right \}$\\
    \textbf{Output}: Subgraph $\tilde{G}$.
    \begin{algorithmic}[1]
        \STATE initialize $\tilde{G}\_list$ as an empty list;
        \STATE Put $G$ into $\phi(\cdot)$ to get its contribution $I({G})$ (Equation \ref{equ9});
        \STATE Put $v^{'}_i$ into $\phi(\cdot)$ to get its contribution $I({v^{'}_i})$, where $v^{'}_i \in (GB_1,  GB_2, \cdots GB_p)$; and put $e^{'}_j$ into $\phi(\cdot)$ to get its contribution $I({e^{'}_j})$, where $e^{'}_j \in (e_1, e_2, \dots, e_q)$ (Equation \ref{equ10}, \ref{equ11});
        \STATE Compare $I({G})$ with $I({v^{'}_i})$ and $I({e^{'}_j})$:
        \IF {$I({v^{'}_i}) > I({G})$}
        \STATE add $v^{'}_i$ to $\tilde{G}\_list$.
        \ENDIF
        \IF {$I({e^{'}_j}) > I({G})$}
        \STATE add $e^{'}_j$ to $\tilde{G}\_list$.
        \ENDIF
        \RETURN $\tilde{G}\_list$.
    \end{algorithmic}  
\end{algorithm}


\subsection{Time Complexity Analysis}
Let $n$ and $m$ denote the number of nodes and edges in the original graph, respectively. The number of nodes and edges in the corresponding structural graph is $p$ and $q$, respectively, and both are much smaller than $n$ and $m$. The time complexity of SeeExplainer is $O((m+n)logn)$. 

Specifically, we perform a global computation on the graph to calculate the degree of each node in the graph, with a time complexity of $O(m+n)$. Then, we perform a coarse-splitting strategy to the original graph to generate $\sqrt{n}$ granular-balls. In this step, $\sqrt{n}$ nodes with the highest degrees are selected as center nodes, and the remaining nodes are assigned using a breadth-first search (BFS) algorithm, resulting in a time complexity of $O(m+n)$. Next, we treat each granular-ball as a parent-ball and recursively split until the algorithm converges, with a time complexity of $O((m+n)logn)$. The time complexity for computing the quality of a granular-ball is $O(p+q)$.

\section{Experiments}
\label{sec_experimental}
To comprehensively evaluate SeeExplainer, we formulate four research questions:
\begin{itemize}
    \item \textbf{RQ1 (Fidelity)}: Can SeeExplainer achieve high fidelity in interpretability?
    \item \textbf{RQ2 (Stability)}: Can the explanatory subgraphs be stably generated? Specifically, at different sparsity levels, the overall fidelity is expected to remain similar.
    \item  \textbf{RQ3 (Ablation Studies)}: How do the key components of SeeExplainer affect its performance?
    \item \textbf{RQ4 (Case Study)}: How does SeeExplainer perform in visualization?
\end{itemize}

\begin{table*}[htbp]
\centering
\footnotesize
\caption{The fidelity comparison between SeeExplainer and baselines on benchmark datasets with GIN and GCN (\%). Higher values indicate better fidelity. For clarity, the best result is highlighted in \textbf{bold} and the second-best result is represented by \underline{underline}.} 
\setlength{\tabcolsep}{0.9mm}
\begin{tabular}{l|c|cccccccccc}
\toprule
GNNs&    Method   & MUTAG            &Mutagenicity      &NCI1              & IMDB-BINARY      & ENZYMES          &PROTEINS          &NCI109            &DHFR              &BZR               &DD           \\
\midrule
    &  PGExplainer&18.284            &7.802             &-0.817            &14.082            &5.064             &13.350            &6.788             &0.820             &0.374             &25.019       \\
    &  DeepLIFT   &30.168            &19.742            &12.179            &4.008             &9.869             &12.790            &10.416            &8.042             &1.576             &-52.911       \\
    & GNNExpaliner&1.076             &5.549             &4.358             &\underline{14.870}&10.244            &15.742            &1.931             &2.770             &8.243             &29.511       \\
 GIN& GraphLime   &\underline{41.270}&11.882            &-5.028            &-4.118            &\underline{39.576}&\underline{30.368}&7.380             &-1.658            &-2.060            &47.270       \\
    &  GradCAM    &12.251            &16.283            &3.512             &-4.038            &0.023             &-5.163            &5.253             &7.816             &15.741            &69.570       \\
    &   Eig-Search&13.409            &\underline{55.563}&\underline{41.870}&11.861            &26.591            &17.124            &\underline{52.972}&\underline{62.048}&\underline{42.258}&\underline{72.127}    \\
    &SeeExplainer &\textbf{64.193}   &\textbf{78.639}   &\textbf{69.842}   &\textbf{53.566}   &\textbf{69.190}   &\textbf{58.437}   &\textbf{68.281}   &\textbf{76.214}   &\textbf{54.947}   &\textbf{95.681}        \\
	\hline
    &  PGExplainer&20.262            &1.331             &3.958             &10.465            &1.176             &11.819            &-0.289            &-4.426             &8.045            &13.639       \\
    &  DeepLIFT   &1.922             &12.625            &3.767             &11.048            &6.267             &13.723            &2.946             &4.224              &-0.683           &20.700       \\
    & GNNExpaliner&1.958             &12.720            &3.618             &11.130            &6.028             &13.700            &3.006             &4.240              &-0.206           &20.054       \\
 GCN& GraphLime   &\underline{38.920}&2.638             &2.868             &-10.792           &15.704            &25.400            &-6.772            &-9.316             &-3.514           &33.710       \\
    &  GradCAM    &2.926             &18.683            &9.068             &3.765             &-0.910            &3.562             &5.165             &20.479             &22.289           &66.919       \\
    &   Eig-Search &6.746            &\underline{57.726}&\underline{57.582}&\underline{24.091}&\underline{18.469}&\underline{28.485}&\underline{36.753}&\underline{54.393} &\underline{66.141}&\underline{88.519}    \\
    &SeeExplainer &\textbf{63.305}   &\textbf{87.700}   &\textbf{82.457}   &\textbf{48.297}   &\textbf{39.052}   &\textbf{62.700}   &\textbf{54.132}   &\textbf{55.280}    &\textbf{75.923}   &\textbf{96.111}        \\
\bottomrule
\end{tabular}
\label{tab3}
\end{table*}

\subsection{Experimental Setup}
\label{}
\noindent \textbf{Datasets:} We evaluate our method on several standard graph classification datasets, including MUTAG \cite{doi:10.1021/jm00106a046}, Mutagenicity \cite{doi:10.1021/jm040835a}, NCI1 \cite{wale2008comparison}, IMDB-BINARY \cite{cai2018simple}, ENZYMES \cite{borgwardt2005protein}, PROTEINS \cite{blondel2008fast}, NCI109 \cite{morris2020weisfeiler}, DHFR \cite{sutherland2003spline}, BZR \cite{vincent2021online}, and DD \cite{morris2020tudataset}. Table \ref{tab1} lists the basic information of datasets.

\begin{table}[ht]
\centering
\footnotesize
\caption{Dataset Information.}
\setlength{\tabcolsep}{1.6mm}
\begin{tabular}{l|ccc}
\toprule
         Datasets    & Graphs   & Avg. Nodes  & Avg. Edges\\
\midrule
		MUTAG        &   188    & 17.93       & 19.79  \\
		  Mutagenicity &   4337   & 30.32       & 30.77  \\
		NCI1         &   4110   & 29.87       & 32.30  \\
		PROTEINS     &   1113   & 39.06       & 72.82  \\
		IMDB-BINARY  &   1000   & 19.77       & 96.53  \\
		NCI109       &   4127   & 29.68       & 32.13  \\
		DHFR         &   756    & 42.43       & 44.54  \\
		BZR          &   405    & 35.75       & 38.36  \\
		DD           &   1178   & 284.32      & 715.66 \\
		ENZYMES      &   600    & 32.63       & 62.14  \\
\bottomrule
\end{tabular}
\label{tab1}
\end{table}

\begin{table}[ht]
\centering
\caption{Benchmark model information. BL, BH, and ACC indicate the best number of layers, the best hidden units, and the best accuracy, respectively.}
\setlength{\tabcolsep}{1.6mm}
\begin{tabular}{l|ccc|ccc}
\toprule
\multirow{2}[4]{*}{Datasets} & \multicolumn{3}{c|}{GIN}  & \multicolumn{3}{c}{GCN} \\
\cmidrule{2-7} & BL  & BH  & ACC           & BL  & BH  & ACC       \\
\midrule
    MUTAG        & 4   & 64  & 0.867 ± 0.083 & 3   & 128 & 0.792 ± 0.124 \\
    Mutagenicity & 3   & 64  & 0.813 ± 0.021 & 3   & 64  & 0.807 ± 0.019   \\
       NCI1      & 3   & 64  & 0.784 ± 0.015 & 3   & 64  & 0.743 ± 0.042   \\
    IMDB-BINARY  & 4   & 128 & 0.750 ± 0.034 & 2   & 64  & 0.748 ± 0.046   \\
     ENZYMES     & 2   & 128 & 0.400 ± 0.057 & 3   & 128 & 0.335 ± 0.100   \\
    PROTEINS     & 4   & 128 & 0.732 ± 0.044 & 3   & 128 & 0.734 ± 0.037   \\
      NCI109     & 4   & 64  & 0.766 ± 0.024 & 3   & 32  & 0.727 ± 0.030   \\
        DHFR     & 4   & 64  & 0.803 ± 0.059 & 3   & 128 & 0.734 ± 0.074   \\
         BZR     & 4   & 32  & 0.840 ± 0.023 & 5   & 128 & 0.852 ± 0.021   \\
          DD     & 1   & 128 & 0.727 ± 0.032 & 3   & 128 & 0.735 ± 0.029   \\
\bottomrule
\end{tabular}
\label{tab2}
\end{table}

\noindent \textbf{Experimental Settings:} In our evaluation, we consider two variants of GNNs, namely Graph Isomorphism Networks (GIN) \cite{xu2019powerfulgraphneuralnetworks} and Graph Convolutional Networks (GCN) \cite{DBLP:journals/corr/KipfW16}. For the training of these GNNs, we use grid search to adjust the hyperparameters and find the best combination. We set the epochs to $100$, the batch size to $128$, the learning rate to $0.01$, the learning rate decay factor to $0.5$, and the learning rate decay step size to $50$. The network hierarchy range is $[1,2,3,4,5]$, and the hidden layer range is $[16,32,64,128]$. Each dataset is split into 80\% training, 10\% testing, and 10\% validation sets. Each model is run 20 times, and the model with the highest test accuracy is our benchmark model. All experiments are conducted on a Xeon(R) Gold 5218 CPU @ 2.30GHz with four Tesla V100 GPUs. The key software environment includes CUDA 11.3, Python 3.9.18, Pytorch 1.11.0, and NetworkX 3.2.1. The detailed information of the model is shown in Table \ref{tab2}.

\noindent\textbf{Baselines.} We extensively compared our method with the most widely used and the state-of-the-art explain algorithms. The comparison methods including PGExplainer \cite{NEURIPS2020_e37b08dd}, DeepLIFT \cite{DeepLIFT}, GNNExplainer \cite{ying2019gnnexplainer}, GraphLime \cite{huang2022graphlime}, GradCAM \cite{Pope_2019_CVPR}, and Eig-Search \cite{lu2024eigsearchgeneratingedgeinducedsubgraphs}.\\

\noindent\textbf{Metrics.} In this paper, we employ standard metrics of fidelity as evaluation criteria, where higher values indicate superior performance. $Fidelity^+$ measures the change in model prediction when the explanatory subgraphs $\tilde{G}$ is removed, while $Fidelity^-$ measures the change when only $\tilde{G}$ is retained. $Fidelity$ is used to quantify the overall change in model prediction. By evaluating $Fidelity^+$ and $Fidelity^-$, we can obtain a comprehensive understanding of explanation accuracy at different sparsity levels $s_k$, thereby capturing the contribution of different input features. Specifically, 
    \begin{equation}
        Fidelity^{+}(G, s_k)=\phi(G, s_k) -\phi(G \setminus \tilde{G}, s_k),
    \end{equation}
    \begin{equation}
        Fidelity^{-}(G, s_k)=\phi(G, s_k) -\phi(\tilde{G}, s_k),
    \end{equation}
    \begin{equation}
        Fidelity(G, s_k) = Fidelity^{+}(G, s_k) - Fidelity^{-}(G, s_k).
    \end{equation}
Note that the fidelity reported in our experimental results is the average of $Fidelity(G, s_k)$ across different sparsity levels $s_k$. 

Stability is defined as the difference between the average fidelity across all sparsity levels and the fidelity at a specific sparsity level. A smaller difference indicates stronger stability. Formally,
    \begin{equation}
        S(s_k) = \left |\frac{1}{u} \sum_{k=1}^{u} Fidelity(G, s_k) - Fidelity(G, s_k) \right |,
    \end{equation}
where $u$ is the number of sparsity levels.

\begin{table*}[htbp]
\centering
\footnotesize
\caption{The stability comparison between SeeExplainer and baselines on the benchmark datasets using GIN and GCN. Lower values indicate better stability. For clarity, the best result is highlighted in \textbf{bold} and the second-best result is represented by \underline{underline}.}
\setlength{\tabcolsep}{0.9mm}
\begin{tabular}{l|c|cccccccccc}
\toprule
GNNs&    Method   & MUTAG             &Mutagenicity       &NCI1               & IMDB-BINARY       & ENZYMES           &PROTEINS           &NCI109             &DHFR               &BZR                &DD                   \\
\midrule
    &  PGExplainer&1.6E-01            &7.4E-02            &3.9E-02            &8.5E-02            &8.7E-02            &6.3E-02            &\underline{6.5E-03}&4.6E-02            &1.2E-02            &1.6E-01       \\
    &  DeepLIFT   &1.0E-01            &5.8E-02            &4.0E-02            &9.1E-02            &7.3E-02            &6.8E-02            &7.4E-02            &\underline{3.1E-02}&1.5E-02            &5.9E-02       \\
    & GNNExpaliner&6.1E-02            &\underline{4.8E-02}&3.4E-02            &8.6E-02            &8.6E-02            &9.4E-02            &2.3E-02            &\underline{3.1E-02}&1.2E-01            &1.7E-01       \\
 GIN& GraphLime   &\underline{3.0E-02}&7.5E-02            &\underline{3.0E-02}&\underline{5.6E-02}&\underline{1.4E-02}&\underline{3.9E-03}&1.1E-02            &3.2E-02            &\underline{4.3E-03}&\underline{4.6E-03}\\
    &  GradCAM    &1.3E-01            &9.6E-02            &4.7E-02            &1.1E-01            &8.6E-02            &8.6E-02            &2.4E-02            &9.6E-02            &7.0E-02            &9.7E-02       \\
    &   Eig-Search&1.5E-01            &1.3E-01            &1.2E-01            &1.3E-01            &1.1E-01            &1.2E-01            &1.0E-01            &1.4E-01            &7.2E-02            &1.6E-02               \\
    &\textbf{SeeExplainer}&\textbf{1.9E-10}   &\textbf{8.0E-11}   &\textbf{1.2E-10}   &\textbf{2.4E-10}   &\textbf{6.1E-10}   &\textbf{2.4E-10}   &\textbf{1.9E-10}   &\textbf{3.2E-10}   &\textbf{4.3E-10}   &\textbf{1.0E-09}      \\
	\hline
    &  PGExplainer&1.1E-01            &5.4E-02            &2.9E-02            &6.9E-02            &4.5E-02            &6.2E-02            &2.4E-02            &\underline{3.4E-02}&1.0E-01            &1.0E-01       \\
    &  DeepLIFT   &3.4E-02            &8.4E-02            &3.4E-02            &6.2E-02            &3.5E-02            &7.5E-02            &2.9E-02            &3.5E-02            &1.1E-0             &1.4E-01       \\
    & GNNExpaliner&3.5E-02            &8.6E-02            &3.5E-02            &6.4E-02            &3.9E-02            &7.8E-02            &2.9E-02            &4.0E-02            &\underline{1.4E-02}&1.4E-01       \\
 GCN& GraphLime   &\underline{1.2E-02}&\underline{4.6E-03}&\underline{1.9E-02}&\underline{5.9E-02}&\underline{5.1E-03}&\underline{7.2E-04}&\underline{3.0E-03}&5.6E-02            &\underline{1.4E-02}&\underline{1.5E-02}\\
    &  GradCAM    &4.7E-02            &1.2E-01            &5.7E-02            &7.0E-02            &4.8E-02            &7.4E-02            &5.6E-02            &3.8E-02            &1.8E-01            &9.7E-02       \\
    &   Eig-Search&5.7E-02            &1.7E-01            &1.3E-01            &9.8E-02            &6.7E-02            &1.4E-01            &1.0E-01            &4.4E-02            &7.1E-02            &1.7E-02               \\
    &\textbf{SeeExplainer}&\textbf{6.8E-10}   &\textbf{8.6E-11}   &\textbf{9.0E-11}   &\textbf{2.7E-10}   &\textbf{1.3E-09}   &\textbf{1.9E-10}   &\textbf{1.0E-10}   &\textbf{2.8E-10}   &\textbf{5.0E-10}   &\textbf{1.2E-08}      \\
\bottomrule
\end{tabular}
\label{tab4}
\end{table*}

\subsection{Fidelity (RQ1)}
A good explainer should be able to generate accurate explanations. To verify the effectiveness of SeeExplainer, we conduct experiments on ten real-world datasets. Table \ref{tab3} shows the average fidelity results on the datasets (MUTAG, Mutagenicity, NCI1, IMDB-BINARY, ENZYMES, PROTEINS, NCI109, DHFR, BZR, and DD) with different graph neural networks (GIN and GCN) at different sparsity levels (0.5, 0.6, 0.7, 0.8, and 0.9). It is evident from Table \ref{tab3} that SeeExplainer achieves the best results compared with baselines across all datasets, regardless of GIN or GCN. \textbf{These results indicate that SeeExplainer effectively captures the synergistic effects among edges during the structuring stage of the graph, thereby improving fidelity.}

\begin{figure}[ht]
    \centering 
    \includegraphics[width=4.2cm,height=2.8593cm]{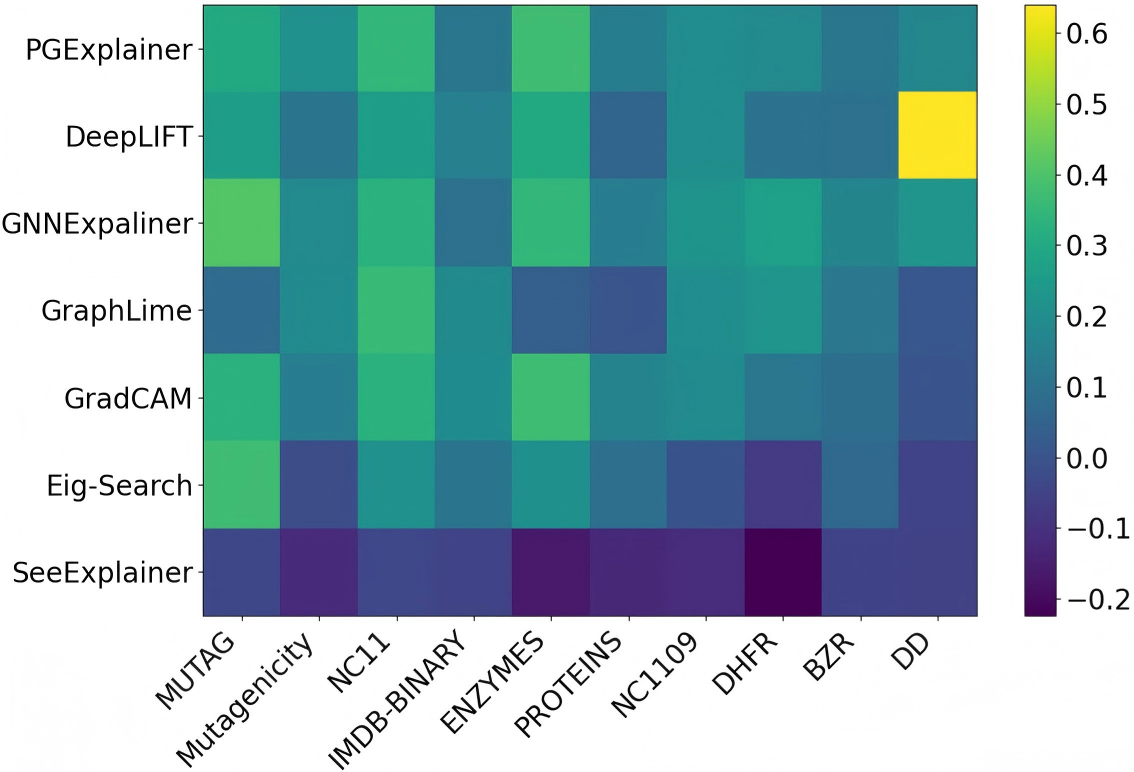}\label{fidelity-gin}
    \includegraphics[width=4.2cm,height=2.8593cm]{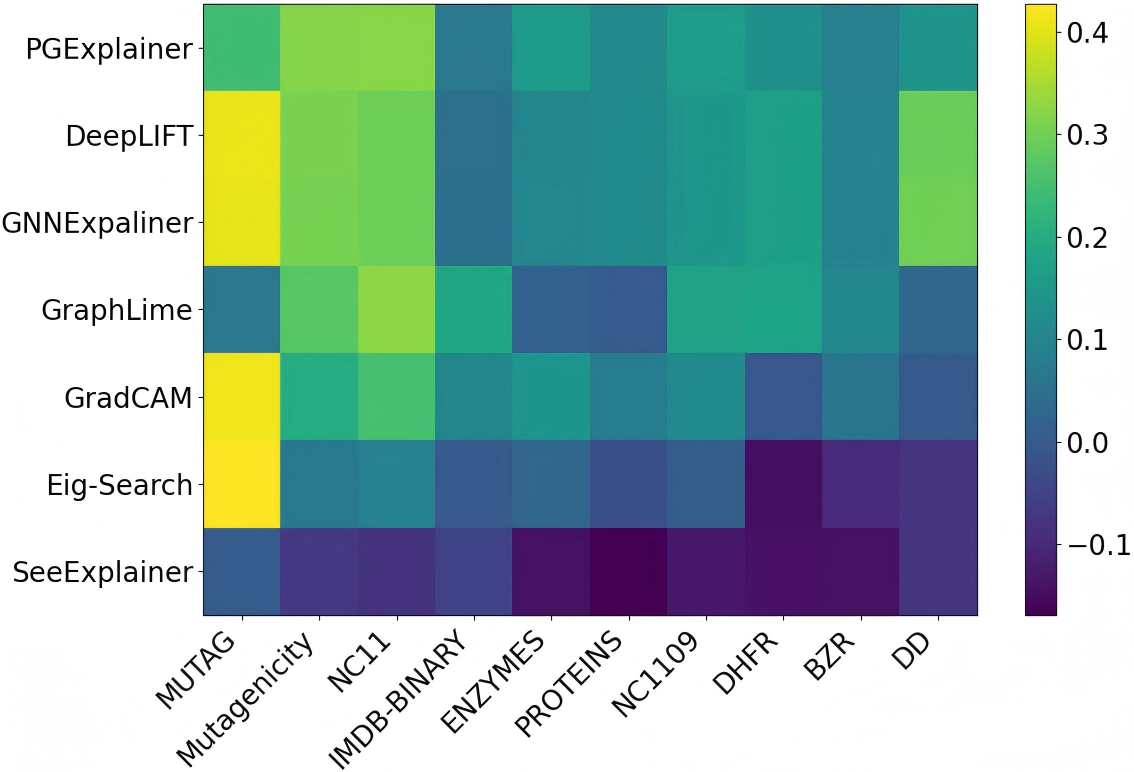}\label{fidelity-gcn}
    \hspace{0.00001cm}
    \includegraphics[width=4.2cm,height=2.8593cm]{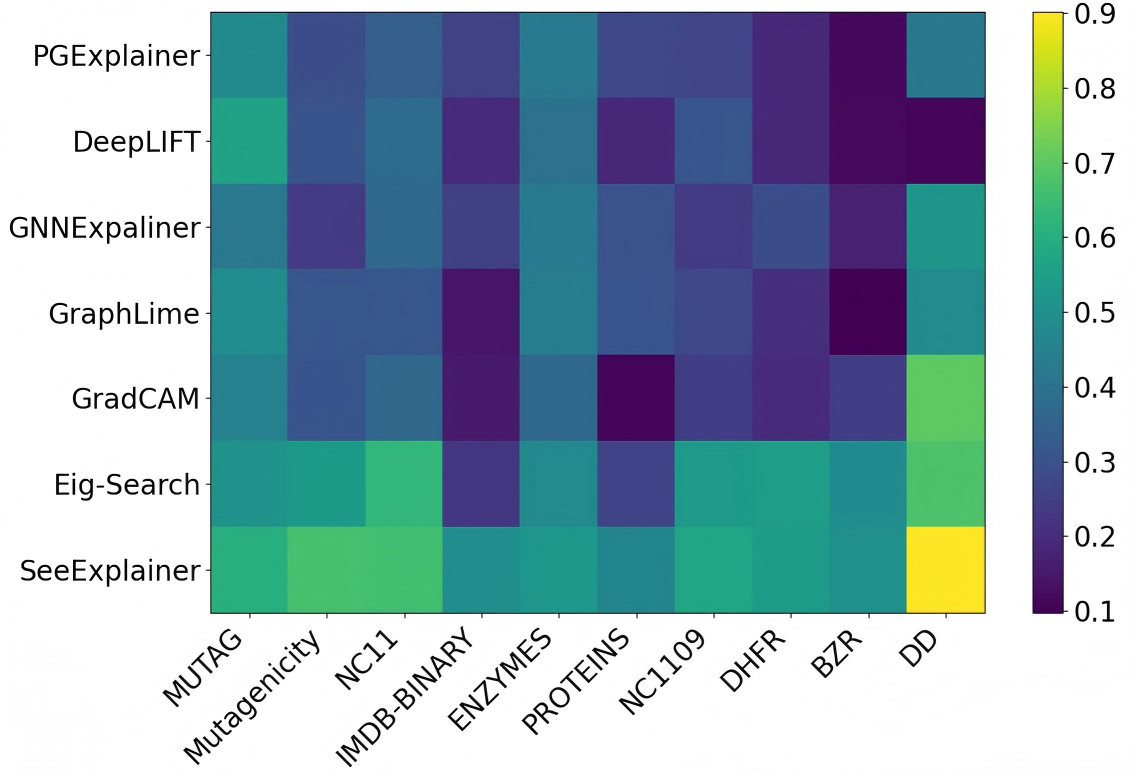}\label{fidelity+gin}
    \includegraphics[width=4.2cm,height=2.8593cm]{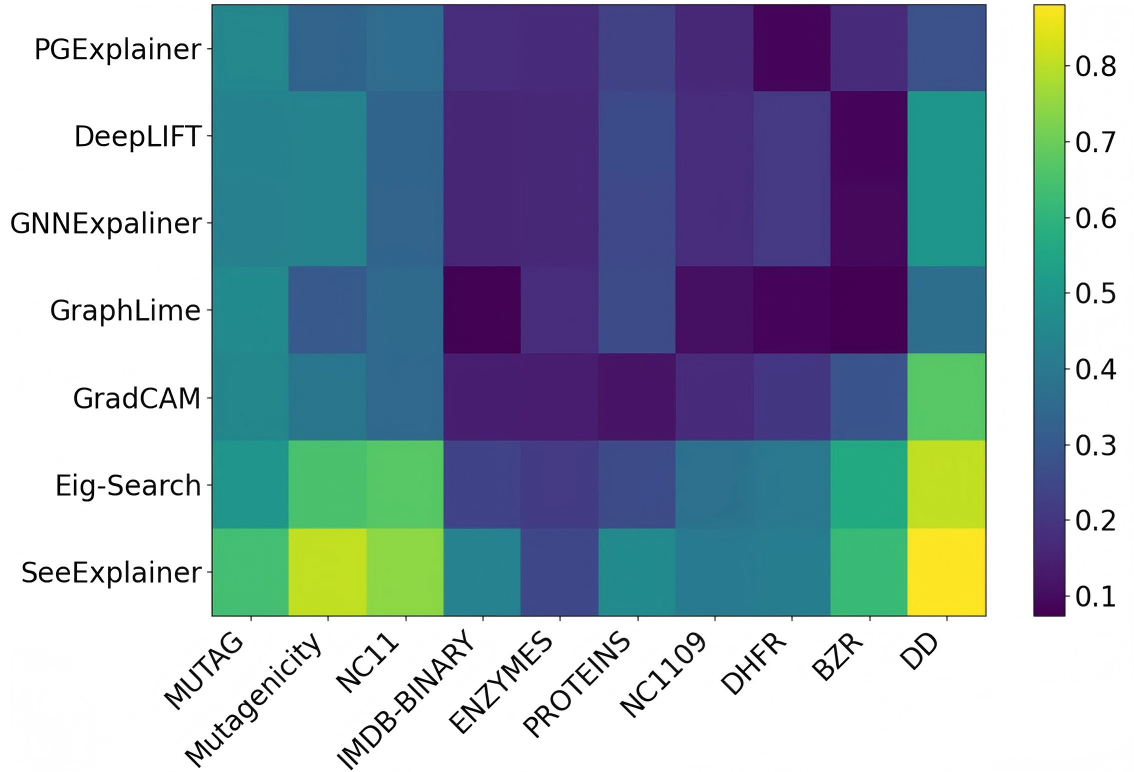}\label{fidelity+gcn}
    \caption{Heatmaps of fidelity$^-$ and fidelity$^+$ for GIN and GCN.}
    \label{fig3}
\end{figure}

Since fidelity is jointly determined by fidelity$^+$ and fidelity$^-$, we present heatmaps of the fidelity$^+$ and fidelity$^-$ values across all datasets under different graph neural networks to better illustrate the effectiveness of SeeExplainer. As shown in Fig. \ref{fig3}, the first row depicts the fidelity$^-$ values for all datasets using GIN or GCN, where a lower fidelity$^-$ indicates better performance. The experimental results show that SeeExplainer achieves the lowest values in most cases. The second row presents the fidelity$^+$ values under GIN or GCN, where a higher fidelity$^+$ indicates better performance. As illustrated in the figure, SeeExplainer consistently attains the highest fidelity$^+$ values in most cases. Consequently, SeeExplainer achieves the best overall performance. Although some baselines have achieve relatively high fidelity$^+$, their correspondingly high fidelity$^-$ values result in suboptimal overall performance. These results indicate that SeeExplainer more effectively captures truly explanatory subgraphs by retaining high-value components while removing low-value ones. It should be noted that the graphs in the DD dataset are very large, and their labels are determined by the overall topological structure. This indicates that SeeExplainer has a significant advantage in capturing and characterizing graph structures.

\subsection{Stability (RQ2)}
In this section, we conduct experiments to evaluate the stability of baselines and SeeExplainer. Stability quantifies the robustness of explanations under different sparsity levels. The average difference in fidelity reflects the degree of fluctuation in the explanations. Accordingly, we evaluate fidelity at different sparsity levels and compute the average difference between the fidelity at each sparsity level and the overall average fidelity across all sparsity levels. Smaller variations indicate that the explainer produces more consistent and stable explanations. The experimental results are reported in Table \ref{tab4}, which shows that SeeExplainer also achieves the best stability performance across all datasets, with values close to $0$. \textbf{The reason is that SeeExplainer generates explanatory subgraphs directly from the structural graph and is therefore unaffected by parameter settings.} In contrast, our baseline method Eig-Search produces explanations that depend on the input parameter $k$, which leads to substantial variation in its output.

\begin{table*}[htbp]
\centering
\footnotesize
\caption{Ablation results of different initialization methods for structural graph generation.
}
\setlength{\tabcolsep}{0.9mm}
\begin{tabular}{l|c|cccccccccc}
\toprule
GNNs&    Variants   & MUTAG             &Mutagenicity       &NCI1               & IMDB-BINARY       & ENZYMES           &PROTEINS           &NCI109             &DHFR               &BZR                &DD                   \\
\midrule
\multirow{2}{*}{GIN} &\multicolumn{1}{l|}{Two}         &62.284         &\textbf{79.479}&\textbf{70.952}&52.558         &68.487          &\textbf{60.289}&67.487         &76.041         &\textbf{55.165}&95.205          \\ 
                     &\multicolumn{1}{l|}{Radical n}   &\textbf{64.193}&78.639        &69.842         &\textbf{53.566}&\textbf{69.190} &58.437         &\textbf{68.281}&\textbf{76.214}&54.947         &\textbf{95.681} \\  
\hline
\multirow{2}{*}{GCN} &\multicolumn{1}{l|}{Two}         &\textbf{63.458}&87.207         &81.348         &47.885         &38.632          &62.656         &53.500         &\textbf{56.760}&75.521         &95.396          \\ 
                     &\multicolumn{1}{l|}{Radical n}   &63.305         &\textbf{87.700}&\textbf{82.457}&\textbf{48.297}&\textbf{39.052} &\textbf{62.700}&\textbf{54.132}&55.280         &\textbf{75.923}&\textbf{96.111} \\ 
\bottomrule
\end{tabular}
\label{tab6}
\end{table*}

\begin{figure}[ht]
\centering
\subfigure[]{
\includegraphics[width=7cm,height=3.6448cm]
    {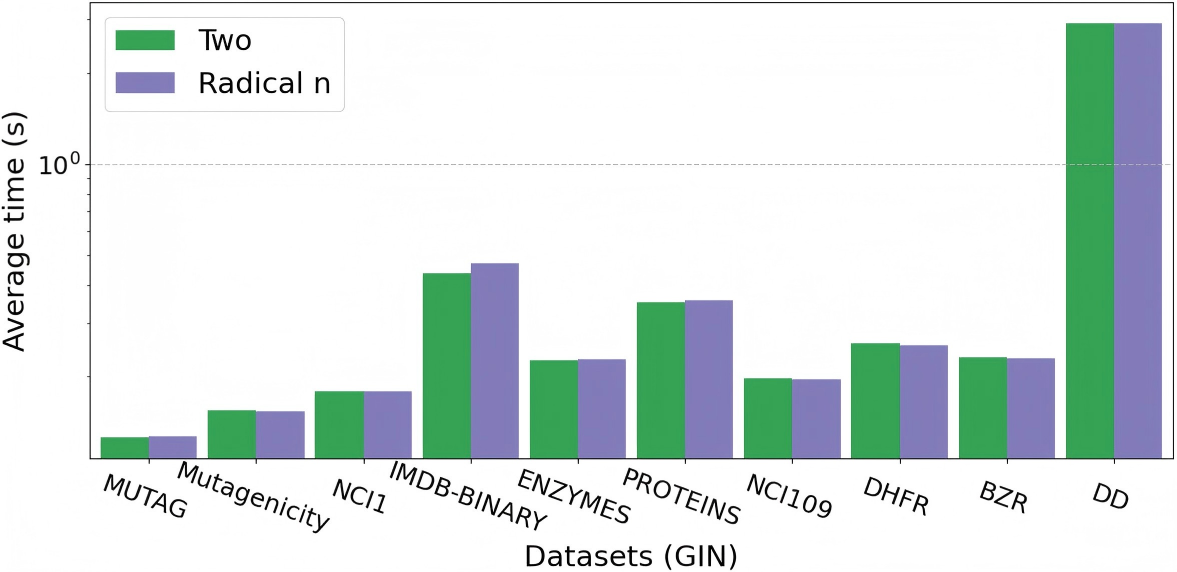}}
\subfigure[]{
\includegraphics[width=7cm,height=3.6448cm]
{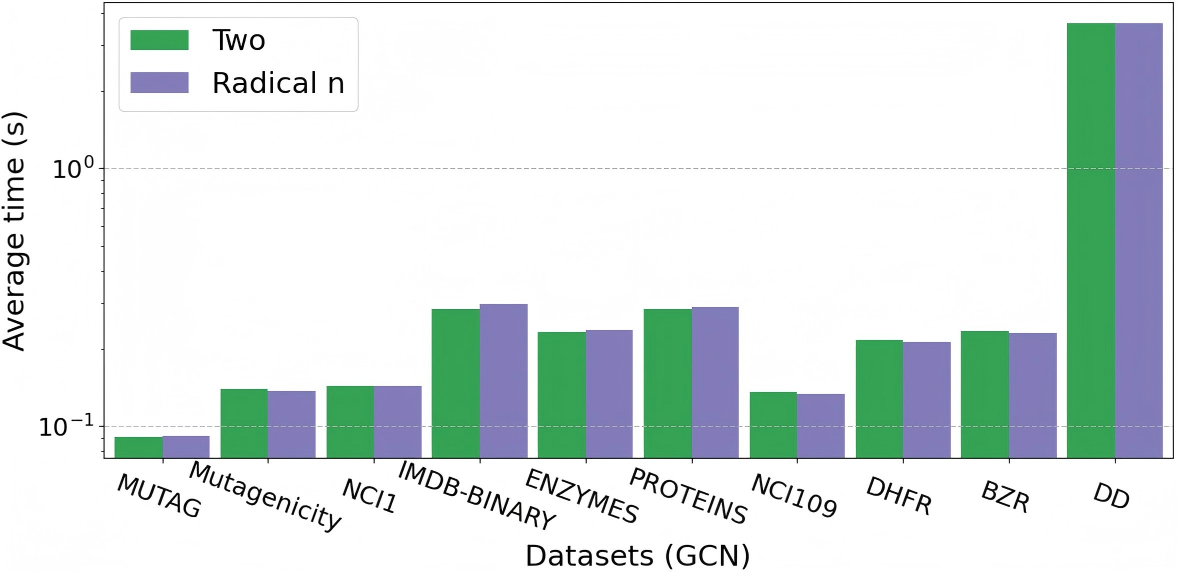}}
\caption{Comparison of time consumption between two and radical n splitting in the first round.}
\label{fig6}
\end{figure}

\subsection{Ablation Studies (RQ3)}
In this section, to better evaluate the role of each module in SeeExplainer, we conducted ablation experiments to assess the performance of the variants, including the different initialization strategies for structural graph generation and the composition information in the explanatory subgraphs. 

\begin{table*}[htbp]
\centering
\footnotesize
\caption{Ablation results of composition information in the explanatory subgraph.}
\setlength{\tabcolsep}{0.9mm}
\begin{tabular}{l|c|cccccccccc}
\toprule
GNNs&    Variants   & MUTAG             &Mutagenicity       &NCI1               & IMDB-BINARY       & ENZYMES           &PROTEINS           &NCI109             &DHFR               &BZR                &DD                   \\
\midrule
\multicolumn{1}{l|}{}    &\multicolumn{1}{l|}{SeeExplainer}      &\textbf{64.193}&\textbf{78.639}     &\textbf{69.842}&\textbf{53.566}  &\textbf{69.190} &\textbf{58.437}&\textbf{68.281}&\textbf{76.214}&\textbf{54.947}&\textbf{95.681} \\ 
\multicolumn{1}{l|}{GIN} &\multicolumn{1}{l|}{-w/o inter-structure edges}        &16.084         &50.612              &28.974         &28.737           &37.220          &27.007         &38.420         &41.986         &25.423         &46.067    \\ 
\multicolumn{1}{l|}{}    &\multicolumn{1}{l|}{-w/o structures}&11.514         &47.116              &37.386         &-9.296           &21.216          &15.727         &50.560         &64.865         &11.207         &40.381    \\ 
\hline
\multicolumn{1}{l|}{}    &\multicolumn{1}{l|}{SeeExplainer}      &\textbf{63.305}&\textbf{87.700}     &\textbf{82.457}&\textbf{48.297}  &\textbf{39.052} &\textbf{62.700}&\textbf{54.132}&\textbf{55.280}&\textbf{75.923}&\textbf{96.111} \\ 
\multicolumn{1}{l|}{GCN} &\multicolumn{1}{l|}{-w/o inter-structure edges}        &9.131          &54.239              &42.785         &19.425           &20.283          &34.871         &31.519         &48.250         &22.115         &78.373    \\ 
\multicolumn{1}{l|}{}    &\multicolumn{1}{l|}{-w/o structures}&2.198          &53.452              &52.735         &3.736            &14.060          &21.768         &33.334         &44.669         &54.350         &59.311    \\ 
\bottomrule
\end{tabular}
\label{tab5}
\end{table*}

\begin{figure*}[ht]
\centering
\includegraphics[width=3.4cm,height=2.2626cm]{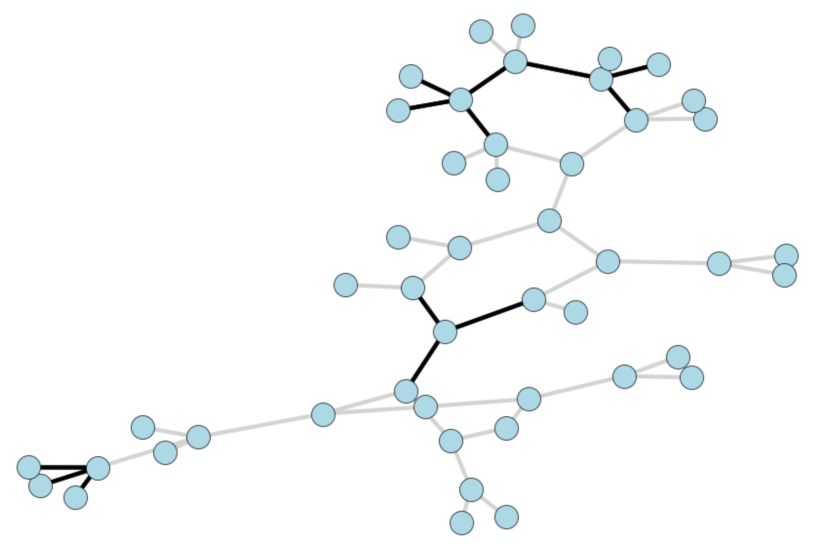}
\hspace{0.7cm}
\includegraphics[width=3.4cm,height=2.2626cm]{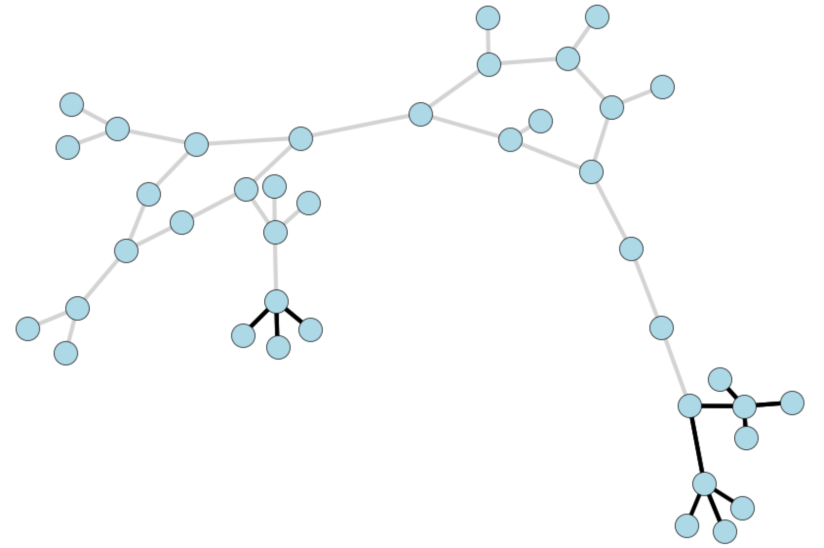}
\hspace{0.7cm}
\includegraphics[width=3.4cm,height=2.2626cm]{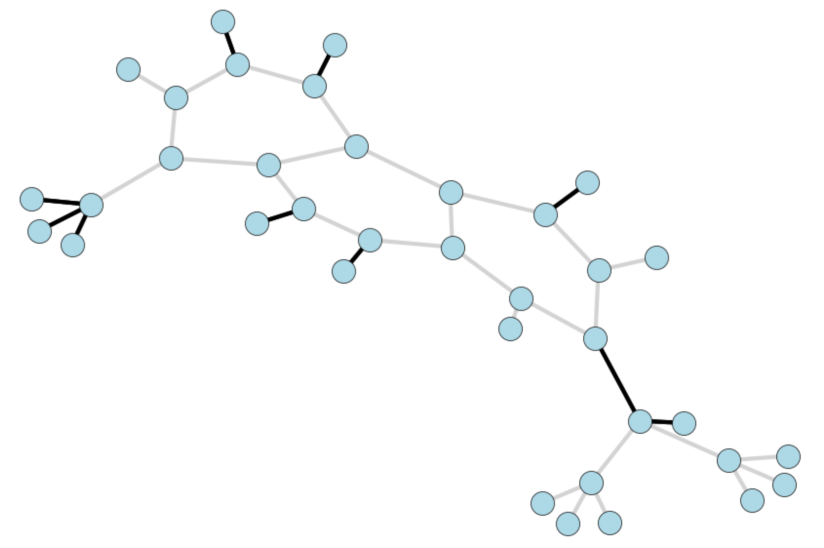}
\hspace{0.7cm}
\includegraphics[width=3.4cm,height=2.2626cm]{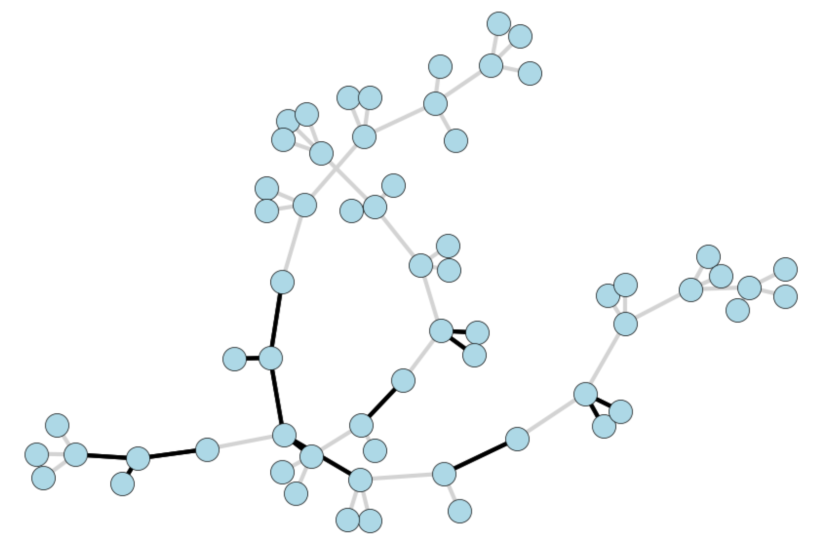}
\\
\includegraphics[width=3.4cm,height=2.2626cm]{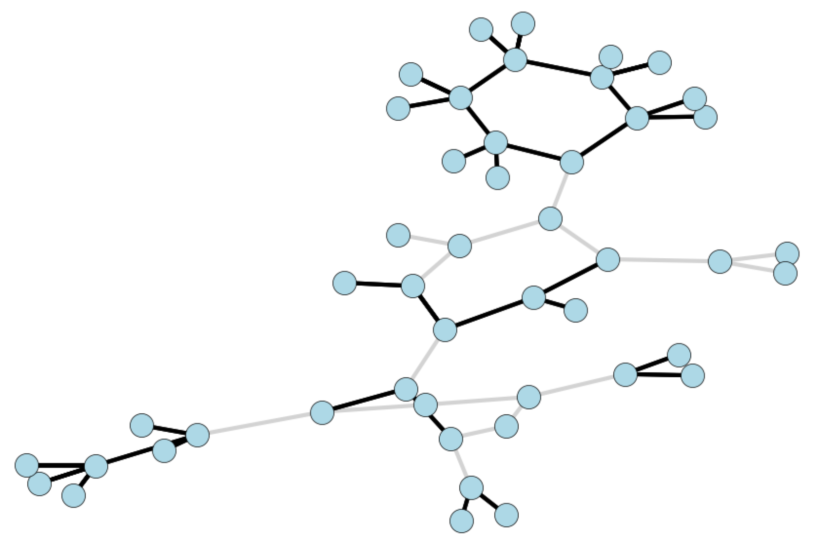}
\hspace{0.7cm}
\includegraphics[width=3.4cm,height=2.2626cm]{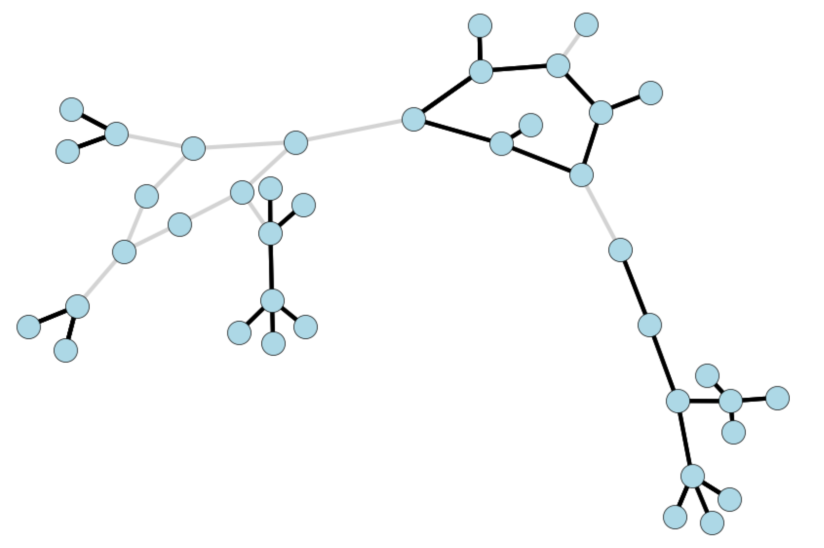}
\hspace{0.7cm}
\includegraphics[width=3.4cm,height=2.2626cm]{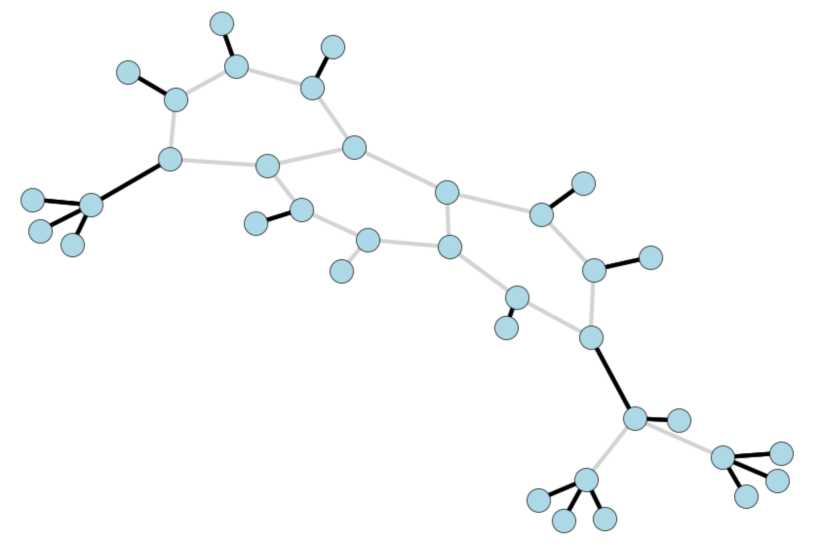}
\hspace{0.7cm}
\includegraphics[width=3.4cm,height=2.2626cm]{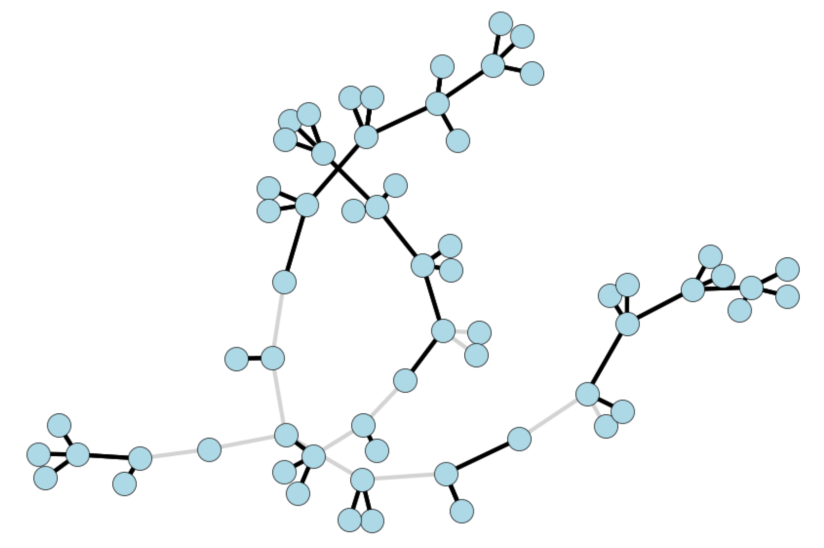}
\\
\includegraphics[width=3.4cm,height=2.2626cm]{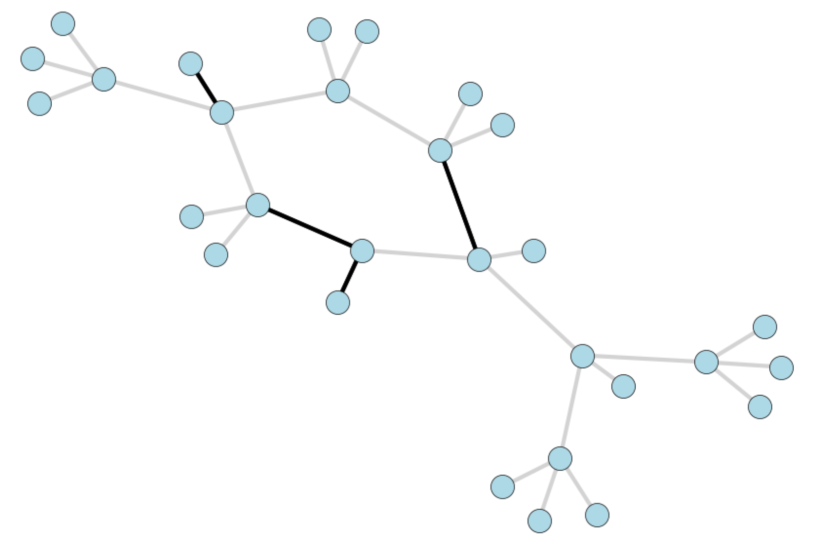}
\hspace{0.7cm}
\includegraphics[width=3.4cm,height=2.2626cm]{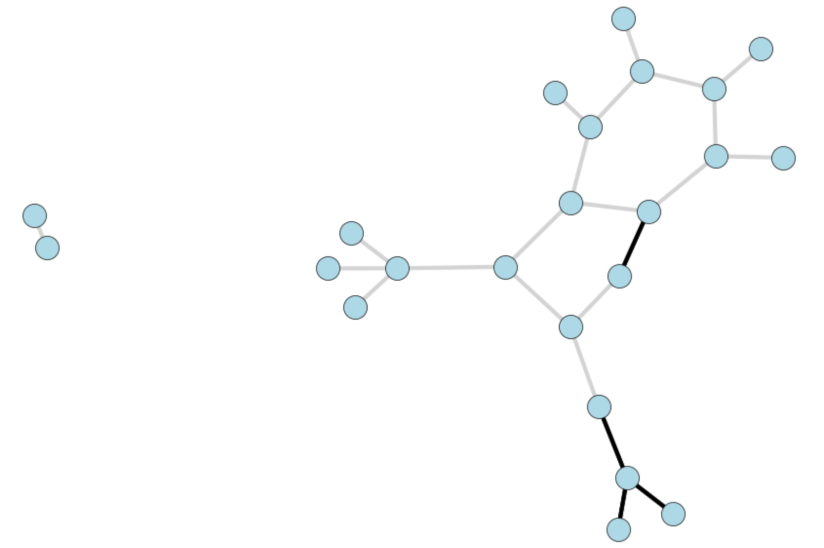}
\hspace{0.7cm}
\includegraphics[width=3.4cm,height=2.2626cm]{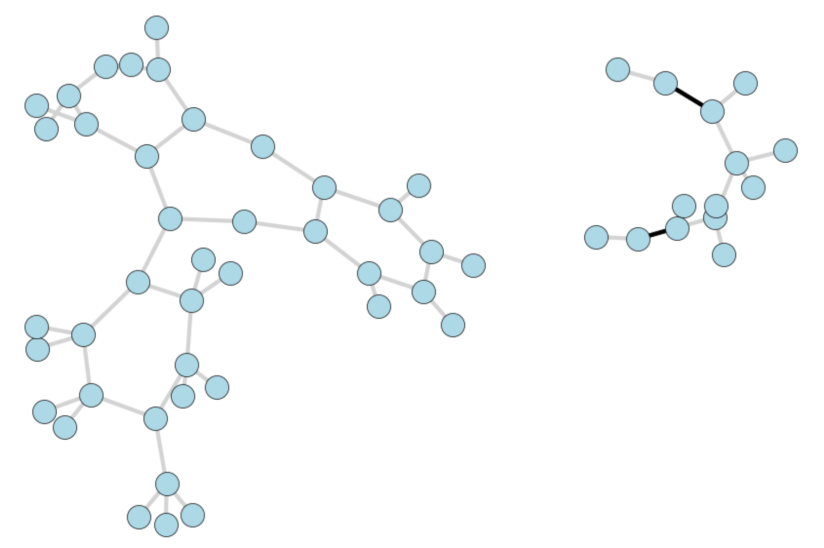}
\hspace{0.7cm}
\includegraphics[width=3.4cm,height=2.2626cm]{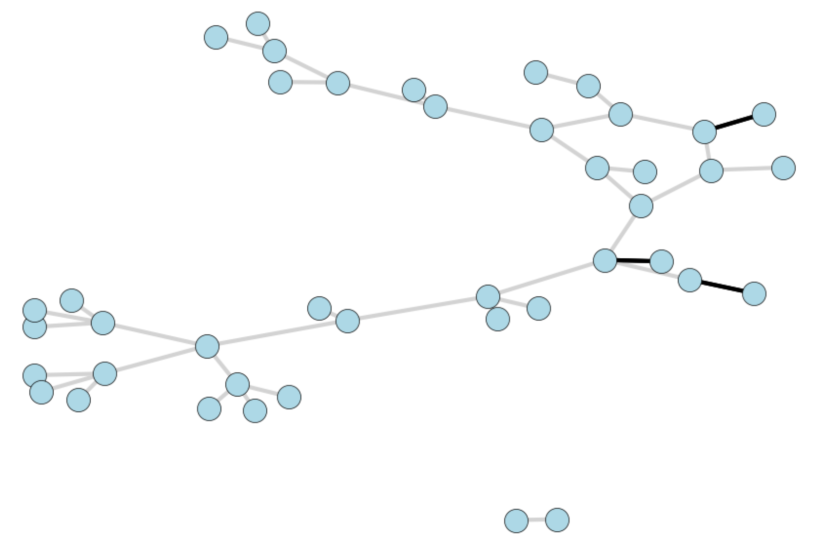}
\\
\includegraphics[width=3.4cm,height=2.2626cm]{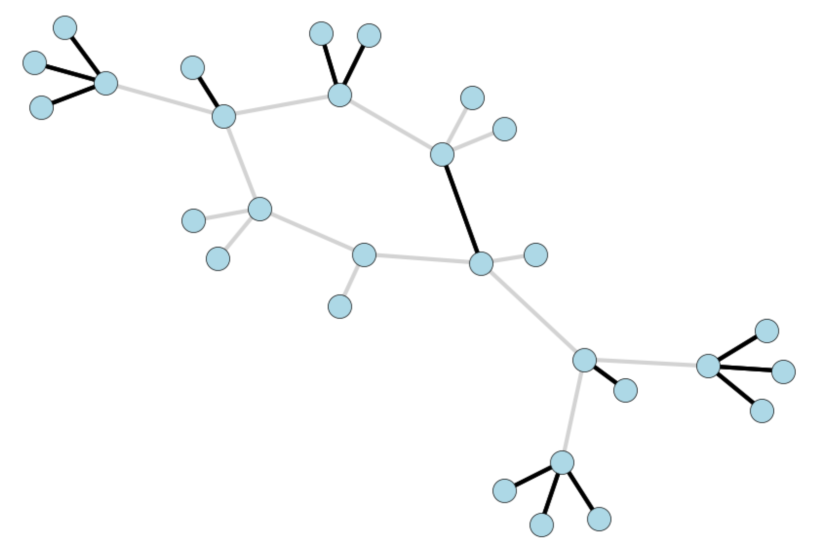}
\hspace{0.7cm}
\includegraphics[width=3.4cm,height=2.2626cm]{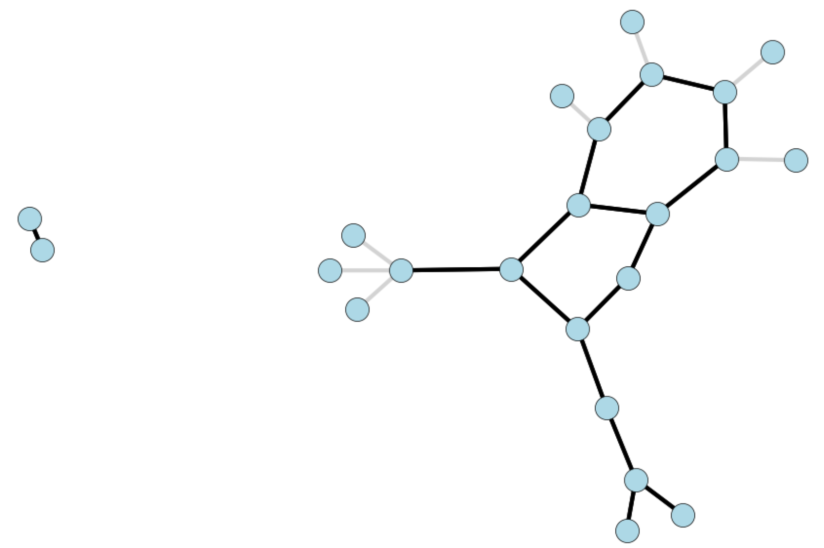}
\hspace{0.7cm}
\includegraphics[width=3.4cm,height=2.2626cm]{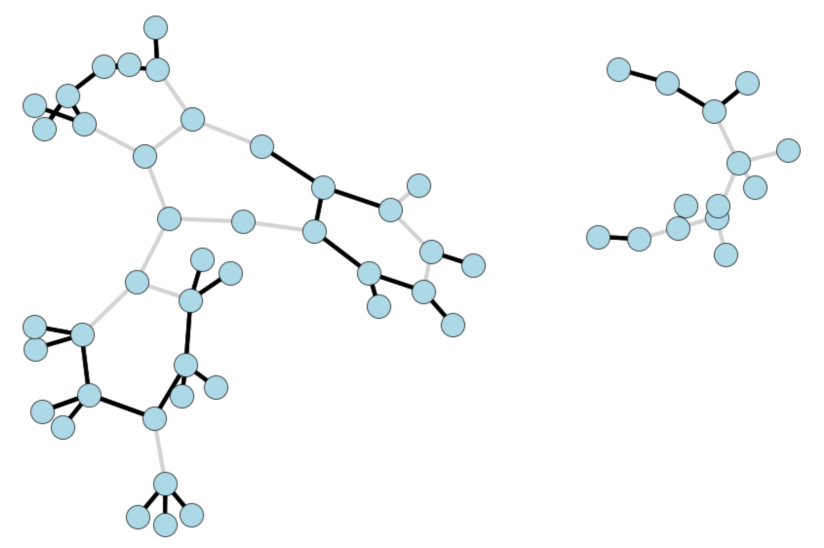}
\hspace{0.7cm}
\includegraphics[width=3.4cm,height=2.2626cm]{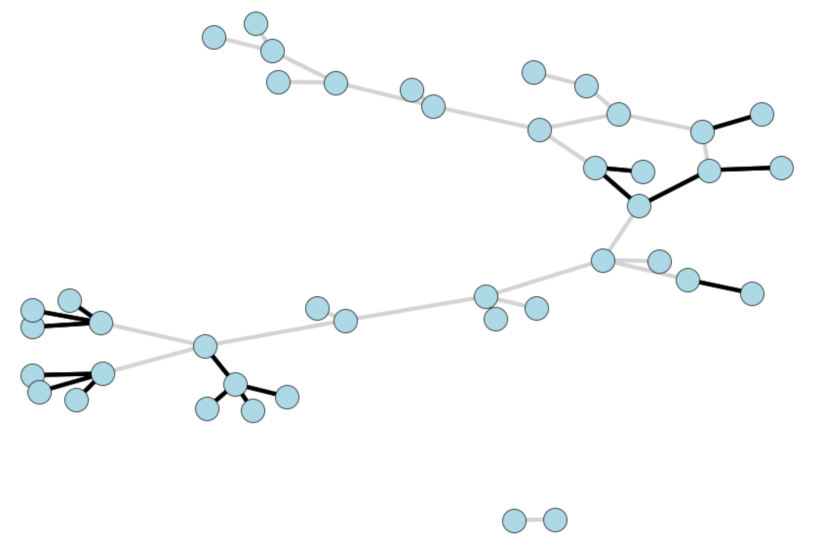}
\\
\caption{Comparison of cases. The first and third rows indicate Eig-Search, and the second and fourth rows represent SeeExplainer. The bold black areas are explanatory subgraphs.}
\label{fig_casestudy}
\end{figure*}

\subsubsection{Initialization methods for structural graph generation}
We analyze the influence of different initialization methods in the first division round of structural graph generation, including two and radical n initial granular-balls. The fidelity comparison results are reported in Table \ref{tab6}, and the comparison of the average time consumption per graph across different datasets is shown in Fig. \ref{fig6}. Based on the experimental results, we can obtain the following observations.

\begin{itemize}
    \item The experimental results in Table \ref{tab6} indicate that in the first division round of the granular-ball, the radical n division method achieves overall superior performance.
    \item As shown in Fig. \ref{fig6}, except for the IMDB-BINARY and PROTEINS datasets, the radical n method is clearly faster in terms of time consumption than the two-initial granular-ball method.
    \item Based on the above observations, we can conclude that adopting the radical n method in the first split is more advantageous. It achieves higher fidelity than the two-initial granular-ball method while also providing faster computation. \textbf{This phenomenon occurs is because initializing radical n granular-balls enables a more comprehensive capture of important graph features and leads to faster convergence.}
\end{itemize}

\subsubsection{Composition information in the explanatory subgraph}
We analyze the influence of the construction information of explanatory subgraphs, including SeeExplainer-w/o inter-structure edges and SeeExplainer-w/o structures. The results are shown in Table \ref{tab5}.

\begin{itemize}
    \item Removing the inter-structure edges in the structural graph significantly reduces the performance of the model, highlighting its contribution. This mechanism is crucial for capturing coarse-grained features of the graph. Without this mechanism, it is difficult for the model to effectively utilize the features of the graph, which emphasizes its necessity in interpretability tasks.
    \item Removing structures from the explanatory subgraph also significantly reduces the performance of the model. This mechanism can effectively capture fine-grained features in the graph, that is, structural information containing the synergistic effects among edges. It emphasizes the key to understanding the composition of explanatory subgraphs.
    \item It is worth noting that on the MUTAG, NCI1, IMDB-BINARY, ENZYMES, PROTEINS, BZR, and DD with GIN, as well as the MUTAG, IMDB-BINARY, ENZYMES, and PROTEINS with GCN, \textbf{the fidelity of SeeExplainer is higher than the sum of SeeExplainer-w/o inter-structure edges and SeeExplainer-w/o structures, indicating that SeeExplainer effectively captures the synergistic effects among structural components.}
    \item The experimental results for subgraphs that contain both the structures in the structural graph and the edges connecting the structures are much greater than those for individual subgraphs. \textbf{The reason is that the explanatory subgraph effectively captures the compensation effect between coarse-grained structures and fine-grained edges, greatly improving fidelity.}
\end{itemize}

\subsection{Case Study (RQ4)}
\label{sec_experimental_case}
We visualize the explainability results of Eig-Search and SeeExplainer on several datasets. When dealing with the MUTAG and Mutagenicity datasets, our goal is to identify chemical groups that are highly correlated with mutagenicity. The NCI1 and NCI109 datasets are used to predict effectiveness against cancer cells, where the objective is to discover chemical substructures associated with anticancer activity. The IMDB-BINARY dataset is used for binary movie genre classification, aiming to identify actor communities and highly connected star clusters. The ENZYMES dataset focuses on enzyme function classification, with the goal of identifying local structures that distinguish different enzymatic functions. The PROTEINS dataset is used to determine whether a protein is an enzyme, aiming to uncover substructures indicative of enzymatic activity. The DHFR dataset is employed to predict whether a compound inhibits DHFR, with the objective of explaining which atomic subgraphs contribute to DHFR inhibition. The BZR dataset is used to predict binding to the benzodiazepine receptor, aiming to identify substructures that determine a molecule’s receptor-binding capability. Finally, the DD dataset is used to distinguish protein structure types. Since DD graphs are extremely large and their labels are determined by global topology, the explanation goal is to identify which macroscopic structures govern protein folding types. The comparison result is shown in Fig. \ref{fig_casestudy}, which indicates that SeeExplainer retains important structure-level information, enabling classifiers to interpret the results of graph classification.

\section{Conclusions and Future Work}
\label{sec_future_work}
In this work, we proposed SeeExplainer, which investigates how to capture the synergistic effects among graph edges in order to generate more faithful explanations for graph neural networks. Firstly, we explain the existence of synergistic effects among edges. Then, inspired by granular-ball computing and graph non-isomorphic decomposition, we represent the original graph as a structural graph to capture edge-level synergy. Finally, explanatory subgraphs are directly generated from the structural graph through simple threshold judgment. Extensive experiments conducted on multiple real-world graph classification datasets verify the effectiveness of SeeExplainer. In future work, we will study SeeExplainer in real application scenarios (such as the medical domain) to evaluate its effectiveness and interpretability in real-world settings.

\end{document}